\documentclass[letterpaper, 10 pt, conference]{ieeeconf}  % Comment this line out if you need a4paper

\IEEEoverridecommandlockouts                              % This command is only needed if 
\usepackage[T1]{fontenc}
\usepackage{amsmath} % assumes amsmath package installed
\usepackage{amssymb}  % assumes amsmath package installed
\usepackage{multirow}
\usepackage{xcolor}
\usepackage{cite}
\usepackage{booktabs}
\usepackage{multirow}
\usepackage{graphicx}
\usepackage{bm}

\usepackage[pagebackref,breaklinks,colorlinks]{hyperref}
\usepackage[capitalize]{cleveref}
\crefname{section}{Sec.}{Secs.}
\Crefname{section}{Section}{Sections}
\Crefname{table}{Table}{Tables}
\crefname{table}{Tab.}{Tabs.}

\newcommand{\ie}{\textit{i.e.}}

\newcommand{\datasetname}{GAPartManip}

\title{\LARGE \bf
GAPartManip: A Large-scale Part-centric Dataset for Material-Agnostic Articulated Object Manipulation
}

\author{
  Wenbo Cui\textsuperscript{* 1,2,7}, 
  Chengyang Zhao\textsuperscript{* 3,4}, 
  Songlin Wei\textsuperscript{* 3,7}, 
  Jiazhao Zhang\textsuperscript{3,7}, 
  Haoran Geng\textsuperscript{3,5}, \\
  Yaran Chen\textsuperscript{6}, 
  Haoran Li\textsuperscript{1,2}
  He Wang\textsuperscript{\dag \ 3,7}\\
\thanks{* Equal Contribution.}
\thanks{\dag \ Corresponding to \href{mailto:hewang@pku.edu.cn}{hewang@pku.edu.cn}.}
\thanks{This work is supported by the Excellent Youth Program of State Key Laboratory of Multimodal Artificial Intelligence Systems.}
\thanks{$^{1}$Institute of Automation, Chinese Academy of Sciences,
        $^{2}$School of Artificial Intelligence, University of Chinese Academy of Sciences,
        $^{3}$CFCS, School of Computer Science, Peking University,
        $^{4}$Carnegie Mellon University,
        $^{5}$University of California, Berkeley,
        $^{6}$Xi'an Jiaotong-Liverpool University,
        $^{7}$Galbot.
}
}

\begin{document}

\maketitle
\thispagestyle{empty}
\pagestyle{empty}

\begin{abstract}

Effectively manipulating articulated objects in household scenarios is a crucial step toward achieving general embodied artificial intelligence. Mainstream research in 3D vision has primarily focused on manipulation through depth perception and pose detection. However, in real-world environments, these methods often face challenges due to imperfect depth perception, such as with transparent lids and reflective handles. Moreover, they generally lack the diversity in part-based interactions required for flexible and adaptable manipulation.
To address these challenges, we introduced a large-scale part-centric dataset for articulated object manipulation that features both photo-realistic material randomization and detailed annotations of part-oriented, scene-level actionable interaction poses. We evaluated the effectiveness of our dataset by integrating it with several state-of-the-art methods for depth estimation and interaction pose prediction. Additionally, we proposed a novel modular framework that delivers superior and robust performance for generalizable articulated object manipulation. Our extensive experiments demonstrate that our dataset significantly improves the performance of depth perception and actionable interaction pose prediction in both simulation and real-world scenarios. 
More information and demos can be found at: 
\href{https://pku-epic.github.io/GAPartManip/}{https://pku-epic.github.io/GAPartManip/}.

\end{abstract}

\section{Introduction}

Articulated objects are ubiquitous in people's daily lives, ranging from tabletop items like microwaves and kitchen pots to larger items like cabinets and washing machines. Unlike simple, single-function rigid objects, articulated objects consist of multiple parts with different functions, featuring varied geometric shapes and kinematic structures, making generalizable perception and manipulation towards them highly non-trivial~\cite{Xiang_2020_SAPIEN}. 
Some existing works tried to simplify this problem by developing intermediate representations to encode the similarities across different objects implicitly, such as affordance~\cite{mo2021where2act, wu2021vatmart, wang2022adaafford,zhao2022dualafford} and motion flow~\cite{eisner2022flowbot3d, zhang2023flowbot++, zhong20233d}, thereby achieving generalization across objects. 
Another series of work~\cite{geng2023gapartnet, geng2023partmanip, geng2024sage} tried to tackle the articulated object perception and manipulation based on a more explicit and fundamental concept called \textit{Generalizable and Actionable Part (GAPart)}, demonstrating more manipulation capabilities attributed to its 7-DoF pose representation compared to value map representation of visual affordance. 
However, we observe that two critical limitations impede their real-world performance. 

\begin{figure}[ht]
    \centering
    \includegraphics[width=0.5\textwidth]{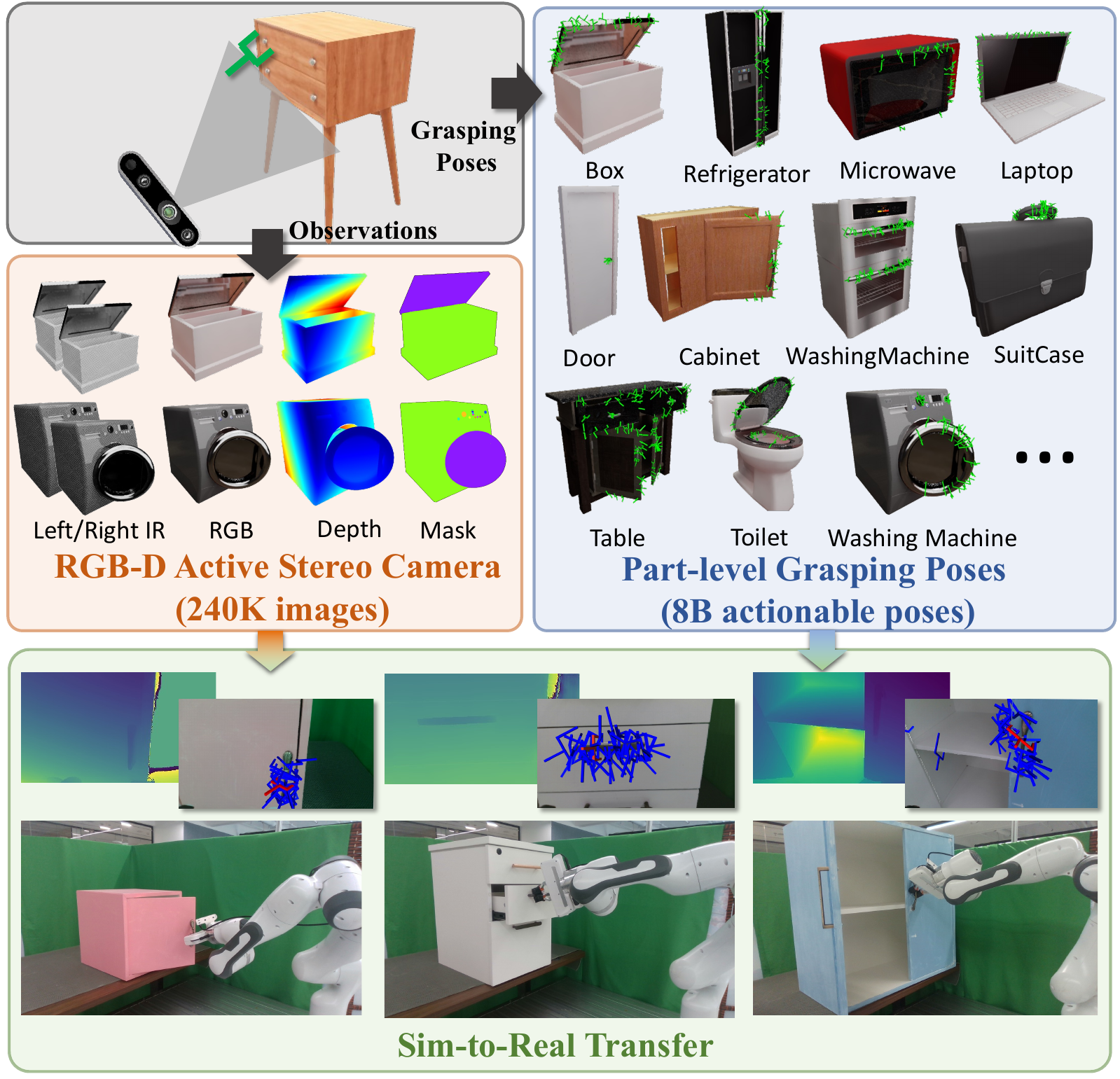}
    \caption{\textbf{GAPartManip}. We introduce a large-scale part-centric dataset for material-agnostic articulated object manipulation. It encompasses 19 common household articulated categories, totaling 918 object instances, 240K photo-realistic rendering images, and 8 billion scene-level actionable interaction poses. GAPartManip enables robust zero-shot sim-to-real transfer for accomplishing articulated object manipulation tasks.}
    \label{fig:teaser}

% \vspace{-5pt}
\end{figure}

Firstly, the material of articulated objects significantly impacts the quality of point cloud data.  
Most existing work relies on point clouds, and these methods struggle due to the sim-to-real gap of depth estimation \cite{geng2023partmanip, geng2023gapartnet, wang2024rpmart, rlafford}. 
Some neural-based stereo-matching depth reconstruction methods are proposed and show some success on rigid objects \cite{wei2024d3roma, shi2024asgrasp}. These methods use neural networks to encode the disparity in stereo infrared (IR) patterns projected by structured light cameras. However, due to the limited diversity in the stereo IR dataset, these methods are constrained to small rigid objects and perform poorly on large articulated objects.

Secondly, there is no method that can predict stable and actionable interaction poses across categories for articulated objects. Some work employs heuristic-based methods \cite{geng2023gapartnet} to interact with articulated objects, but it is limited in diversity and fails to account for the geometric details necessary for robust interactions in real-world settings \cite{wu2021vatmart}. Some methods for rigid object grasping pose prediction can generate stable poses. However, due to the lack of data on articulated objects, it is challenging to discern whether each link can be interacted with independently, resulting in poses that are mostly non-actionable \cite{morlans2023aograsp}. Affordance-based methods \cite{mo2021where2act, rlafford, an2024rgbmanip} receive widespread attention for interacting with articulated objects by generating heatmaps. However, these heatmaps are ambiguous, hard to annotate, and struggle to produce stable interaction poses \cite{wang2024rpmart}.

In this paper, we address these limitations from a data-centric perspective. We introduce~\textit{\datasetname}, a novel large-scale synthetic dataset that features two important aspects: 1) realistic, physics-based IR image rendering for various parts in diverse scenes, and 2) part-oriented actionable interaction pose annotations for a wide range of articulated objects. Our~\datasetname~inherits 918 object instances across 19 categories from the previous GAPartNet dataset\cite{geng2023gapartnet}. By leveraging these assets, we develop a novel data generation pipeline for part manipulation, producing the synthetic data needed to address the previously mentioned limitations. To improve generalizability and mitigate the sim-to-real gap, we incorporate domain randomization techniques \cite{shi2024asgrasp} during data generation, ensuring a diverse range of outputs. In total, our dataset contains approximately 14,000 scene-level samples with 8 billion part-oriented actionable pose annotations, encompassing a wide array of physical materials, object states, and camera perspectives. 

Through training on the proposed dataset, we obtain a depth reconstruction network and an actionable pose prediction network separately to address the two limitations mentioned earlier.
Moreover, we compose these two neural networks modular to a novel articulated object manipulation framework. Through extensive experiments in both simulation and the real world, our method achieves state-of-the-art (SOTA) performance in both individual modular experiments and articulated object manipulation experiments.

To summarize, our main contributions are as follows:

\begin{itemize}

\item We introduce~\datasetname, a novel large-scale dataset with various articulated objects featuring realistic, physics-based rendering and diverse scene-level, part-oriented actionable interaction pose annotations.

\item We propose a novel framework for articulated object manipulation and evaluate each module separately, demonstrating superior effectiveness and robustness compared to baseline methods.

\item We conduct comprehensive experiments in the real world and achieve SOTA performance on articulated object manipulation tasks.

\end{itemize}

\section{Related work}

\begin{figure*}[ht]
    \centering
    \includegraphics[width=0.9\textwidth]{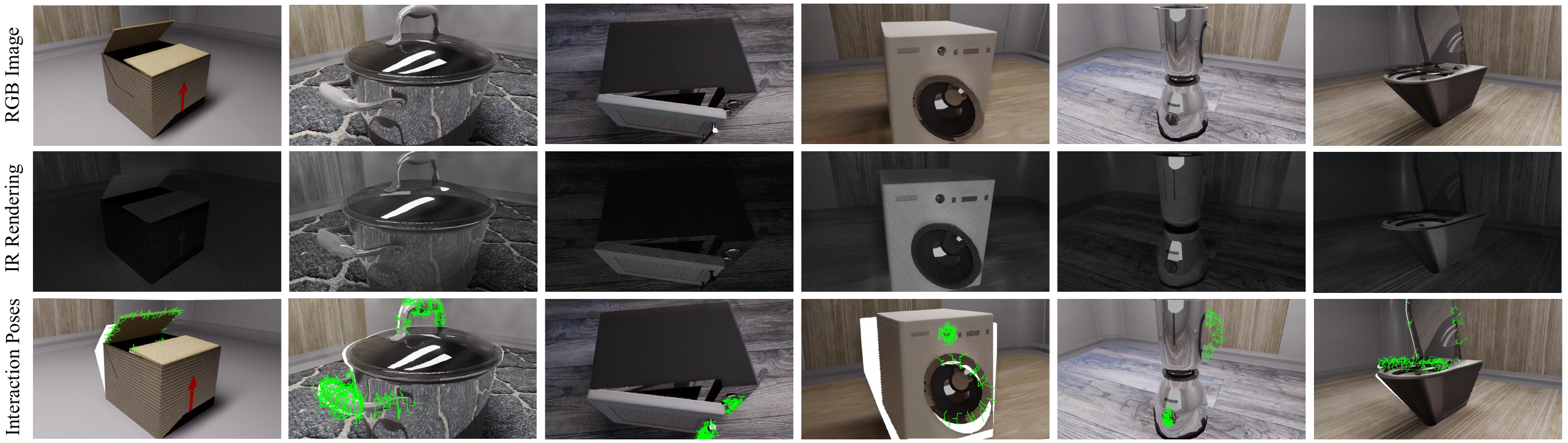}
    \caption{\textbf{Data Examples in~\datasetname.} ~\datasetname~is a novel large-scale synthetic dataset for articulated objects, featuring two important aspects: 1) realistic, physics-based IR rendering for various object materials in diverse scenes, and 2) part-oriented actionable interaction pose annotations for a wide range of articulated objects. Each column shows a data sample. From top to bottom, each row displays the RGB image, the IR image (only the left IR image is shown here), and the scene-level actionable interaction pose annotations.}
    \label{fig:dataset-gallery}
% \vspace{-15pt}
\end{figure*}

\subsection{Articulated Object Dataset}
Articulated object dataset and modeling is a crucial and longstanding research field in 3D vision and robotics, encompassing a wide range of work in perception~\cite{geng2023gapartnet, yi2018deep, deng2024banana, li2020category, liu2023semi, lyu2024scissorbot,zhang2024gamma}, generation~\cite{chen2023urdformer, mu2021sdf, jiang2022ditto, tseng2022cla,luo2024physpartphysicallyplausiblecompletion}, and manipulation~\cite{geng2023gapartnet, lei2023nap, liu2024cage, geng2024sage, geng2022end, geng2023partmanip, gong2023arnold,kuang2024ram}. As to manipulation dataset, GAPartNet\cite{geng2023gapartnet} annotates 7-DoF part pose to manipulate parts. GraspNet\cite{fang2020graspnet} and Contact-Grasp\cite{sundermeyer2021contactgraspnetefficient6dofgrasp} build several datasets, but these datasets all focus on rigid objects, neglecting the kinematic semantics specific to articulated objects. 
Where2Act\cite{mo2021where2act} first introduces a data generation pipeline for articulated objects, which generates data by sampling successful poses in the simulator. AO-Grasp\cite{morlans2023aograsp} leverages a curvature-based sampling method to accelerate data collection efficiency and proposes an 87K dataset for actionable poses. RPMart\cite{wang2024rpmart} manually annotates affordance maps for articulated objects and provides rendering data in SAPIEN\cite{Xiang_2020_SAPIEN}. 
None of the current datasets provide sufficient photo-realistic rendering data to improve the algorithm's capability of perception for articulated objects during sim-to-real, limiting the real-world performance, especially with imperfect point clouds \cite{wang2024rpmart,geng2023partmanip}. Additionally, the data collection processes are inefficient and result in small datasets, hindering the algorithm's generalizability to unseen objects. This work aims to create a large-scale part-centric dataset with diverse photo-realistic rendering and extensive actionable pose data for articulated object manipulation.

\subsection{Articulated Object Manipulation}
Due to complex kinematic structures and geometric shapes, articulated objects present significant challenges in manipulation. Current methods can be broadly categorized into learning-based methods and prediction-planning methods. Learning-based methods, such as reinforcement learning \cite{geng2023partmanip, geng2022end} and imitation learning \cite{gong2023arnold, guhur2023instruction}, require either the realistic and accurate simulation or a large amount of high-quality robot demonstration. However, collecting such data is both impractical and time-consuming, and their sim-to-real performance heavily relies on the quality of the simulation. 
Current prediction-planning methods\cite{mo2021where2act, geng2023gapartnet, geng2024sage,liu2024composablepartbasedmanipulation, ling2024articulatedobjectmanipulationcoarsetofine, kuang2024ram} focus on visual affordance but offer ambiguous interaction poses and struggle to generalize due to the limited data. These methods typically rely on accurate 3D point cloud input, ignoring the impact of object materials on the quality of point clouds. In practice, depth sensors often struggle with challenging materials like glass or metal, missing critical geometric structures on the objects such as handles and lids, significantly limiting the model's sim-to-real performance.

\section{\datasetname~ Dataset}

\subsection{Overview}

We construct a large-scale dataset, \datasetname, to address both depth estimation and actionable interaction pose prediction challenges in articulated object manipulation in real-world scenarios from a data-centric perspective. 
It contains 19 common household articulated categories from GAPartNet, including \textit{Box, Bucket, CoffeeMachine, Dishwasher, Door, KitchenPot, Laptop, Microwave, Oven, Printer, Refrigerator, Safe, StorageFurniture, Suitcase, Table, Toaster, Toilet, TrashCan, and WashingMachine}, comprising a total of 918 object instances after removing problematic assets.

\begin{figure}[ht]
    \centering
    \includegraphics[width=1\linewidth]{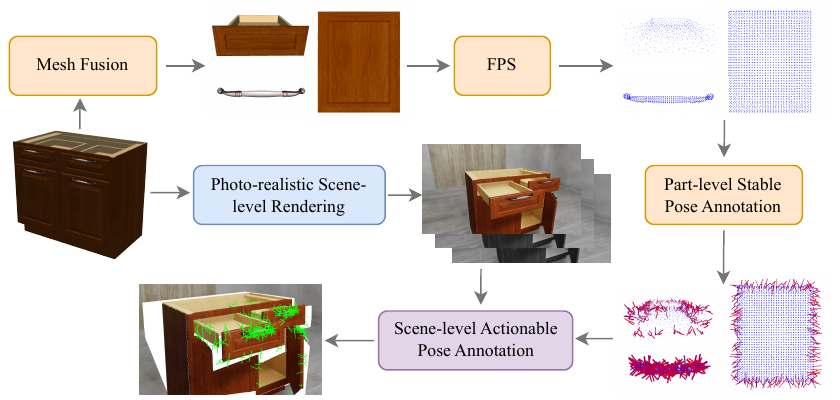}
    \caption{\textbf{Dataset Generation Pipeline.} For scene-level data sample rendering, we input the object asset into our photo-realistic rendering pipeline, generating one RGB image and two IR images (left and right) for each camera perspective. For pose annotation, we begin by performing mesh fusion on each GAPart on the object to establish a one-to-one correspondence between GAParts and meshes. Then, we use FPS to obtain the point cloud for each GAPart, enabling part-level stable interaction pose annotation. These poses are further utilized in scene-level actionable interaction pose annotation for each rendering data sample.}
    \label{fig:dataset-generation}
% \vspace{-10pt}
\end{figure}

We build a photo-realistic rendering pipeline for each asset in indoor scenes.
We render RGB images, IR images, depth maps, and part-level segmentations. 
Additionally, we create high-quality and physics-plausible interaction pose annotations for each part on the articulated object.
Then, we leverage our GPU-accelerated scene-level pose annotation pipeline to generate dense, part-oriented, actionable interaction pose annotations for each rendering data sample.
Our dataset contains over 8 billion actionable poses across 241,680 rendering data samples.
\cref{fig:dataset-gallery} shows examples of data samples from our dataset. Our whole data generation pipeline is illustrated in~\cref{fig:dataset-generation}.

\subsection{Photo-realistic Scene-level Rendering}
Our photo-realistic rendering pipeline is built upon NVIDIA Isaac Sim \cite{isaac-sim}. 
Specifically, we simulate the RGB and IR imaging process of Intel RealSense D415, a widely-used structured light camera for real-world depth estimation in previous research works. 
We replicate the layout of the D415 imaging system consisting of four hardware modules,~\ie, an infrared (IR) projector, an RGB camera, and two IR cameras. 
We also project a similar shadow pattern onto the scenes with D415. 

Inspired by previous works~\cite{dai2022dreds, wei2024d3roma}, we incorporate domain randomization techniques into our rendering pipeline to mimic the IR rendering under various lighting conditions and material properties in the real world. We render each object in 20 different scenes with various domain randomization settings.
Concretely, we randomly vary ambient lighting, background, and object material properties in the scene, generating more diverse data that covers a wider range of real-world imaging conditions. 
we further randomize the ambient light positions and intensities within each scene. 
More importantly, we randomize the parameters of 
all  \textit{diffuse, transparent, specular, and metal} materials for each part corresponding to their semantics.
Finally, we uniformly randomize the joint poses of the object within its joint limits in each scene during the rendering process.

We render the objects and parts from different distances. We render each scene with 5 object-centric camera perspectives for the whole object and 5 part-centric camera perspectives for each part.
To place the object within the camera view,~\ie, the object-centric perspective, the camera is positioned at a latitude of ranged in [10°,60°] and a longitude ranged in [-60°,60°] in the target object frame. 
To capture more fine-grained details of the parts,~\ie, the part-centric perspective, we leverage part pose annotations in GAPartNet and the current joint poses to determine the position and orientation of each part in the scene. 
The camera is then randomly positioned around each part, aiming directly toward the part center.
As a result, the target part occupies the primary area of the image. 
During this process, camera viewpoints are randomly sampled within a latitude range of [0°,60°] and a longitude range of [-75°,75°].

\subsection{GPU-accelerated Scene-level Pose Annotation}

\paragraph{Part-level Stable Pose Annotation}
We employ a pose sampling strategy similar to GraspNet~\cite{fang2020graspnet} to annotate dense and diverse stable interaction poses for each GAPart, based on the original semantic annotations in GAPartNet~\cite{geng2023gapartnet}. 
First, we perform mesh fusion for each part, merging the meshes corresponding to the same part to establish a one-to-one correspondence between parts and meshes. Then, we apply Farthest Point Sampling (FPS) to downsample the mesh of each part, resulting in $N$ candidate points for pose sampling. For each candidate point, we uniformly generate $V \times A \times D$ candidate poses, where $V$ is the number of gripper views distributed uniformly over a spherical surface, $A$ represents the number of in-plane gripper rotations, and $D$ refers to the number of gripper depths. In our case, $N = 512$, $V = 64$, $A = 12$, and $D = 4$.
We follow GraspNet to calculate the pose score based on the antipodal analysis. 

\paragraph{Scene-level Actionable Pose Annotation}
To obtain part-centric interaction poses, We first project the part-level interaction poses into the scene using the part pose annotations, and then filter out unreasonable and unreachable poses. 
More concretely, we classify poses that do not align with single-view partial point clouds as unreasonable.
Meanwhile, we consider poses that cause collisions between the gripper and other parts of the object or the scene as unreachable.

However, such a filtering process is extremely computationally demanding due to the large amounts of points in the scene. 
To accelerate the pose annotation, we implement a CUDA-based optimization for the filtering process.
Our optimization significantly reduces the processing time from 5 minutes to less than 2 seconds for each part, which is nearly a 150-times speed-up. 
As a result, the originally year-long pose annotation process can now be completed within 3 days.

\section{Framework}

\begin{figure*}[ht]
    \centering
    \includegraphics[width=0.95\textwidth, ,height=0.18\textheight]{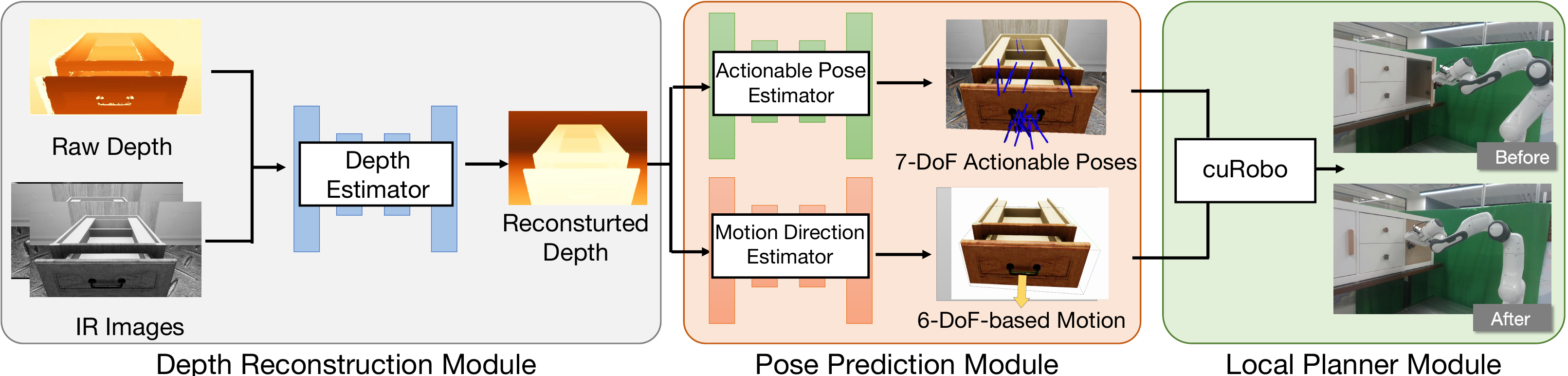}
    % \vspace{-1mm}
    \caption{\textbf{Framework Overview}. Given IR images and raw depth map, the depth reconstruction module first performs depth recovery. Subsequently, the pose prediction module generates a 7-DoF actionable pose and a 6-DoF post-grasping motion for interaction based on the reconstructed depth. Finally, the local planner module carries out the action planning and execution.}
    \label{fig:pipeline}
\end{figure*}

We propose a novel framework to address cross-category articulated object manipulation in real-world settings. 
As illustrated in Fig. \ref{fig:pipeline}, the framework primarily consists of three modules: a depth reconstruction module, a pose prediction module, and a local planner module. 

\subsection{Depth Reconstruction Module}
The input to our system is a single view RGB-D observation including a raw depth $I_d$, a left IR image $I^l_{ir}$, a right IR image $I^r_{ir}$, and an RGB image $I_c$.
The raw sensor depth is often incomplete and even incorrect because transparent and reflective surfaces are inherently ambiguous for structured lights and time-of-flight depth sensors.
To tackle this problem, we leverage diffusion model-based approaches to estimate and restore the incomplete depth for raw sensor outputs.
Specifically, we use D$^3$RoMa \cite{wei2024d3roma} as our depth predictor and fine-tune it on our dataset.

\subsection{Pose Prediction Module}
Different from the 7-DoF grasping pose prediction for rigid object manipulation, here we need to predict both the 7-DoF actionable interaction pose and the 6-DoF interaction motion.

We adapt the SOTA method EconomicGrasp \cite{wu2024economicframework6dofgrasp} as our actionable pose estimator dubbed Part-aware EcoGrasp.
To precisely estimate the part-centric interaction pose, we propose to estimate \textit{actioness} instead of \textit{graspness} in contrast to EconomicGrasp.
To annotate the actioness for learning, we first denote the scene as a point cloud $\mathcal{P}=\{p_i\}_{i=1}^N$ with $N$ points. 
Then for each point $p_i$, we uniformly discretize its sphere space into $V$ approaching directions $\{v_j\}_{j=1}^V$.
For each view $v_j$ of point $p_i$, we generate $L$ actionable pose candidates $A_k^{i,j} \in \textnormal{SE(3)} $ indexed by $k \in \{1,2,\cdots,L\}$ by grid sampling along gripper depths and in-plane rotation angles respectively. 
We employ the antipodal analysis \cite{fang2020graspnet} to calculate the quality score $q_k^{i,j} \in [0, 1.2]$. 
Next, We define an actionable label $c^i_{act} \in \{0,1\}$ for each point, indicating whether this point is on an actionable part.
We also define a scene-level collision label $c_k^{i,j}\in \{0,1\}$ for each pose indicating whether this pose will cause collision.
Finally, the point-wise actioness score $s^P_i$ and the view-wise actioness score $s^V_{i,j}$  are defined as: 

\begin{equation}
% \scriptsize
s^P_i = \frac{1}{\textstyle\displaystyle\sum_{j,k}\left| A^{i,j}_k\right |} {c_{act}^i\textstyle \displaystyle\sum_{j,k}\textbf{1}\left ( q_k^{i,j}>T\right ) c_k^{i,j}},
\end{equation}

\begin{equation}
% \scriptsize
s^V_{i,j}  = \frac{1}{\textstyle \displaystyle\sum_{k}\left | A_k^{i,j}\right |}{  c_{act}^i\textstyle\displaystyle\sum_{k}\textbf{1} \left ( q_k^{i,j}>T\right ) c_k^{i,j}},
\end{equation}
where $T$ is a pre-defined threshold to filter out inferior-quality poses.
We train the Part-aware EcoGrasp following the original setting~\cite{wu2024economicframework6dofgrasp}.

We utilize the pre-trained GAPartNet \cite{geng2023gapartnet} to predict the interaction motion, which specifies the post-grasping movement of the end-effector after grasping the actionable part for interaction.

\subsection{Local Planner Module}

We use CuRobo \cite{curobo} as our motion planner. 
The planner optimizes motion trajectories to the actionable poses estimated by the pose prediction module, computes robot joint angles through inverse kinematics, and drives the robot to execute trajectory actions through joint control. After reaching the target pose, the robot executes grasping for interaction and subsequently executes the post-grasping motion from the pose prediction module to complete the interaction.

\section{Experiments}

We conduct experiments for each module. 
The depth estimation and actionable pose prediction experiments are conducted to illustrate the significance of our dataset in articulated object manipulation tasks. 
Meanwhile, real-world experiments are carried out to compare the performance of our framework with existing methods.
We also performed ablation studies for each module.

\subsection{Depth Estimation Experiments}

In this section, we evaluate different depth estimation methods with our~\datasetname~dataset to demonstrate the effectiveness of our dataset for improving articulated object depth estimation in both simulation and the real world.

\noindent{\textbf{Data Preparation.}}
We split the dataset into training and testing sets using an approximate 8:2 ratio. To maintain comprehensive coverage, each object category is split carefully, ensuring that both the training and testing sets include samples from all categories. 
Additionally, we make sure that samples rendered from the same object category are assigned exclusively to either the training or testing set.
We compare our method with following baselines:
\begin{itemize}[leftmargin=10pt]

\item \textbf{SGM}~\cite{hirschmuller2008sgm} is one of the most widely-used traditional algorithm for dense binocular stereo matching.

\item \textbf{RAFT-Stereo (RS)}~\cite{lipson2021raftstereo} is a learning-based binocular stereo matching architecture built upon the dense optical flow estimation framework RAFT~\cite{teed2020raft}, using an iterative update strategy to recursively refine the disparity map.

\item \textbf{D$^3$RoMa (DR)}~\cite{wei2024d3roma} is a SOTA, learning-based stereo depth estimation framework based on the diffusion model. It excels at restoring noisy depth maps, especially for transparent and specular surfaces.

\end{itemize}

\noindent{\textbf{Evaluation Metrics.}} 
We evaluate the estimated disparity and depth using the following metrics:

\begin{itemize}[leftmargin=10pt]

\item \textbf{EPE}: Mean absolute difference between the ground truth and the estimated disparity map across all pixels.

\item \textbf{RMSE}: Root mean square of depth errors across all pixels.

\item \textbf{MAE}: Mean absolute depth error across all pixels.

\item \textbf{REL}: Mean relative depth error across all pixels.

\item $\bm{\delta_i}$: Percentage of pixels satisfying $\max\left(\frac{d}{\hat{d}}, \frac{\hat{d}}{d}\right) < \delta_i$. $d$ denotes the estimated depth. $\hat{d}$ denotes the ground truth.

\end{itemize}

\begin{table}[ht]
\scriptsize
\centering
\caption{\textbf{Quantitative Results for Depth Estimation in Simulation} 
}
\label{tab:depth_sim}
% \vspace{-15pt}
\begin{center}
\setlength{\tabcolsep}{3pt}
\renewcommand{\arraystretch}{1}
\begin{tabular}{l|cccccccc}
\toprule
Methods & EPE $\downarrow$ & RMSE $\downarrow$ & REL $\downarrow$ & MAE $\downarrow$ & $\delta_{1.05} \uparrow $ & $\delta_{1.10} \uparrow $ &  $\delta_{1.25} \uparrow $\\
\midrule

SGM~\cite{hirschmuller2008sgm} & 6.82 & 1.623 & 0.561 & 0.794 & 34.71 & 38.94 & 46.27 \\
\midrule

RS~\cite{lipson2021raftstereo} & 5.28 & 1.497 & 0.506 & 0.618 & 36.82 & 41.05 & 49.92 \\
\midrule

DR~\cite{wei2024d3roma} & 2.82 & 0.732 & 0.268 & 0.317 & 46.22 & 67.62 & 83.09 \\
\midrule

RS*~\cite{lipson2021raftstereo} & 2.79 & 0.798 & 0.247 & 0.309 & 52.83 & 68.30 & 80.15 \\
\midrule

Ours* & \textbf{0.69} & \textbf{0.225} & \textbf{0.041} & \textbf{0.050} & \textbf{86.22} & \textbf{93.45} & \textbf{97.41}\\

\bottomrule
\end{tabular}
\end{center}
* indicates that the method is fine-tuned on the~\datasetname~dataset.
% \vspace{-10pt}
\end{table}

\begin{figure}[ht]
    \centering
    \includegraphics[width=0.45\textwidth]{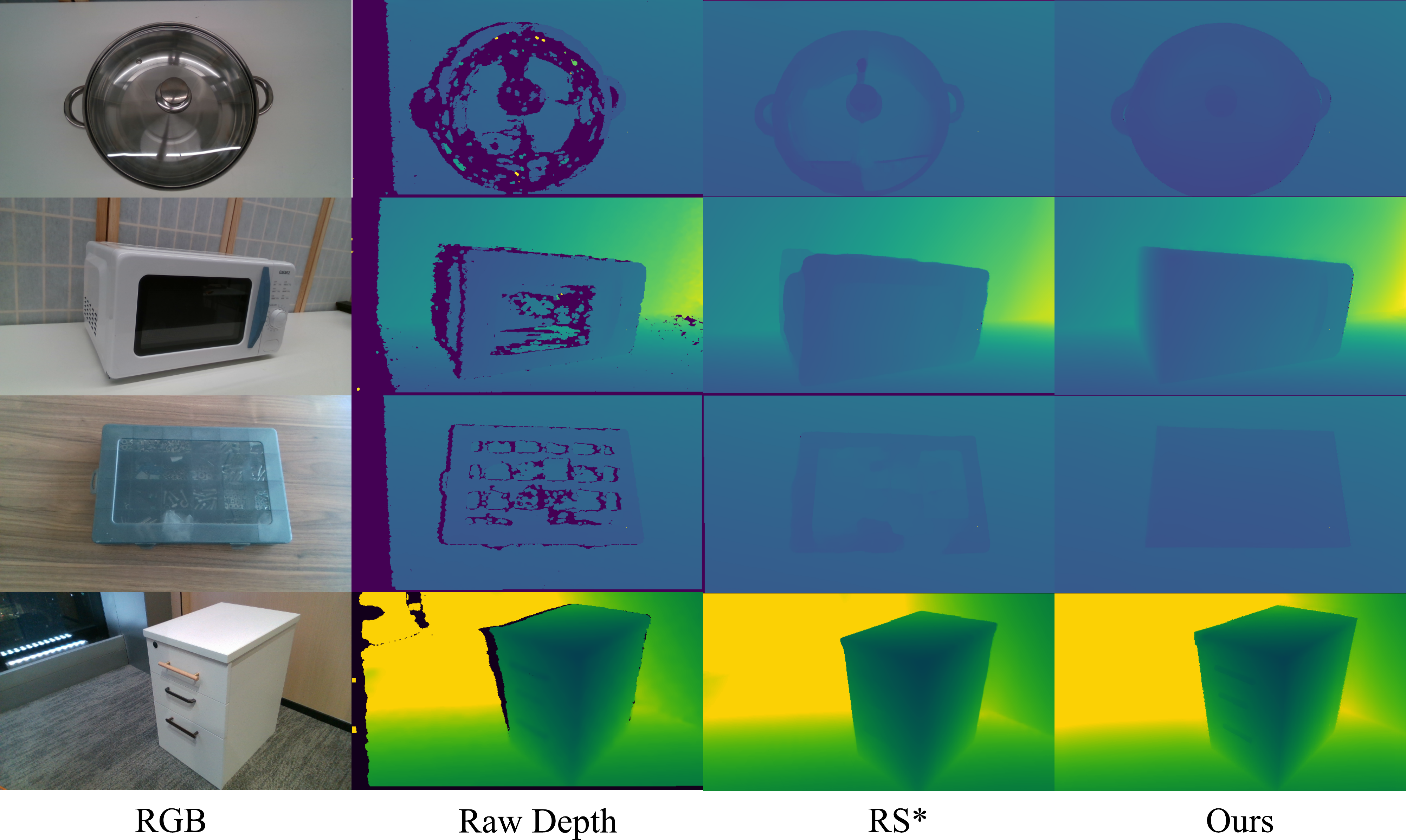}
    \caption{\textbf{Qualitative Results for Depth Estimation in the Real World}. 
    Our refined depth maps are cleaner and more accurate than the ones from the baseline, indicating that our depth reconstruction module is more robust for transparent and translucent lids and small handles.
    Zoom in to better observe small parts like handles and knobs.
    }
    \label{fig:depth_real}
% \vspace{-10pt}
\end{figure}

\noindent{\textbf{Results and Analysis.}}
The quantitative results in simulation are presented in~\cref{tab:depth_sim}. 
The results indicate that the traditional stereo matching algorithm, SGM, struggles in scenes with articulated objects with challenging material characteristics. 
The same observation applies to the pre-trained RAFT-Stereo. Meanwhile, the pre-trained D$^3$RoMa models demonstrate reasonably good stereo depth estimation capabilities in the experiments. 
However, both RAFT-Stereo and D$^3$RoMa are significantly enhanced when fine-tuned on~\datasetname. 
Specifically, RAFT-Stereo achieves a 150\% improvement in MAE compared to its pre-trained version, while our model exhibits a 600\% improvement in MAE, achieving the best performance in the simulation.
As illustrated in~\cref{fig:depth_real}, the fine-tuned models also demonstrate strong depth estimation performance in real-world scenarios. In particular, in real-world environments with challenging materials, as shown in the first three rows of the figure, our model significantly outperforms the fine-tuned RAFT-Stereo and the raw depth, exhibiting noticeably better robustness.
Both simulation and real-world experiments demonstrate the effectiveness of our proposed~\datasetname in substantially improving depth estimation for articulated objects with challenging materials.

\subsection{Actionable Pose Prediction Experiments}\label{sec:pose_experiments}

In this section, we evaluate the impact of our dataset on improving the method for articulated object actionable pose estimation.

\noindent{\textbf{Data Preparation.}}
We split the dataset into training and testing sets using an approximate 7:3 ratio. We further divide the testing set into 3 categories: seen instances, unseen but similar instances, and novel instances.
We compare our methods with the following baselines:

\begin{itemize}[leftmargin=10pt]
\item \textbf{GSNet (GS)} \cite{wang2021gsnet}  is a grasping pose prediction model trained on the GraspNet-1 billion \cite{fang2020graspnet} dataset for rigid objects. We evaluate both the pre-trained model and the fine-tuned model separately.

\item \textbf{Where2Act (WA)}\cite{mo2021where2act} is an affordance-based method for interacting with articulated objects. Unlike the original approach, we do not train separate networks for each task. 
Since Where2Act cannot generate grasping pose prediction, we integrated GSNet, as referenced in \cite{wang2024rpmart}, to enhance Where2Act's capability to align with the experimental setting.

\item \textbf{EconomicGrasp (EG)} \cite{wu2024economicframework6dofgrasp} is also a pose prediction method for rigid objects, which includes an interaction grasp head and composite score estimation to enhance the accuracy of specific grasp prediction.

\end{itemize}

\noindent{\textbf{Evaluation Metrics.} 
Following \cite{fang2020graspnet}, we utilize precision to evaluate the performance of actionable pose estimation:
 
\begin{equation}
% \small
% \vspace{-5pt}
    Precision_\mu=n_{suc_\mu} / n_{grasp}.
% \vspace{-5pt}
\end{equation}

\noindent $Precision_\mu$ represents the ratio of successful interaction pose prediction under the specific friction coefficient $\mu$, where $n_{grasp}$ denotes the number of predicted poses, and $n_{suc_\mu}$ denotes the number of successful grasps under $\mu$. 

\begin{figure}[t]
    \centering
    \includegraphics[width=0.45\textwidth, ]{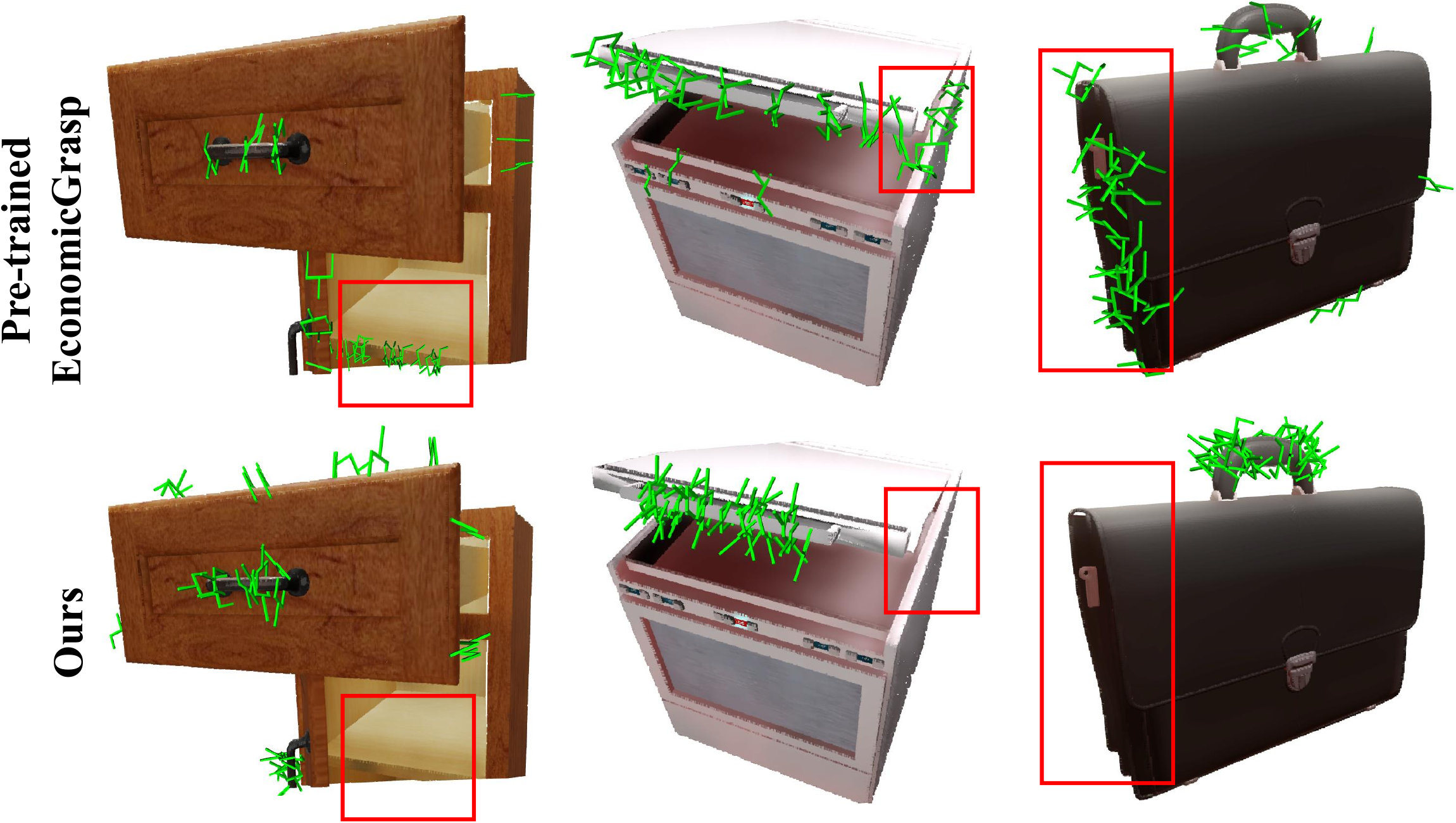}
    \caption{\textbf{Qualitative Comparison of Actionable Pose Prediction in Simulation}.}
    \label{fig:actionable_pose}
    % \vspace{-15pt}
\end{figure}

\noindent{\textbf{Results and Analysis.} 
Our quantitative results in simulation are presented in Tab. \ref{table:pose_estimation}. Even though all fine-tuned on our dataset, both GSNet and our Part-aware EcoGrasp outperform Where2Act, possibly because Where2Act struggles with cross-category and cross-action reasoning. Our model and the fine-tuned GSNet show a substantial improvement in precision compared to the pretrained models. It is evident that our dataset significantly enhances the capability of existing methods in actionable pose estimation for articulated objects. Specifically, our dataset offers strong geometric priors for parts, enabling networks to focus more on the actionable parts rather than the non-actionable ones. For instance, although the pre-trained EconomicGrasp in Fig. \ref{fig:actionable_pose} generates a set of stable grasping poses, it cannot differentiate whether these poses act on actionable parts, meaning they may fail in interacting with articulated objects.

\begin{table}[ht]
\scriptsize
\centering
\begin{center}
\caption{\textbf{Quantitative Results for Actionable Pose Prediction in Simulation} 
}
\setlength{\tabcolsep}{3pt}
\renewcommand{\arraystretch}{1}
\begin{tabular}{cccc|ccc|ccc}
\hline
\toprule
\multirow{2}{*}{Method} & \multicolumn{3}{c|}{Seen}               & \multicolumn{3}{c|}{Unseen}             & \multicolumn{3}{c}{Novel}               \\ \cline{2-10} 
                        & $P$ & $P_{0.8}$ & $P_{0.4}$ & $P$ & $P_{0.8}$ & $P_{0.4}$ & $P$ & $P_{0.4}$ & $P_{0.8}$ \\ \hline
GS\cite{wang2021gsnet}                   &   13.28       &      11.55        &      6.70        &     17.36      &      15.57        &       9.19       &     9.76      &      8.43        &     5.25         \\ 
EG\cite{wu2024economicframework6dofgrasp}                   &   24.72       &      19.65        &      9.97        &     23.91      &      20.29        &       9.90       &     14.56      &      12.02        &     9.23         \\ 

GS*\cite{wang2021gsnet}                   &     25.70      &     20.26         &    9.00          &    25.45       &     20.28         &     9.67         &     23.99      &       20.55       &         11.20     \\ 
WA*\cite{mo2021where2act}               &   14.43        &   12.44           &     6.53     &    11.04       &     7.41         &     2.52         &    4.17       &       1.85       &     0.47         \\ 
Ours*           &     \textbf{55.33}      &       \textbf{51.19}       &        \textbf{30.25}      &     \textbf{56.26}      &     \textbf{53.02}         &     \textbf{32.91}         &    \textbf{41.65}       &      \textbf{39.06}        &        \textbf{23.25}      \\ 
\bottomrule
\end{tabular}
\label{table:pose_estimation}
\end{center}
* indicates that the method is fine-tuned on the~\datasetname~dataset.
\end{table}

\subsection{Real-World experiment}

\begin{figure}[t]
    \centering
    \includegraphics[width=0.9\linewidth]{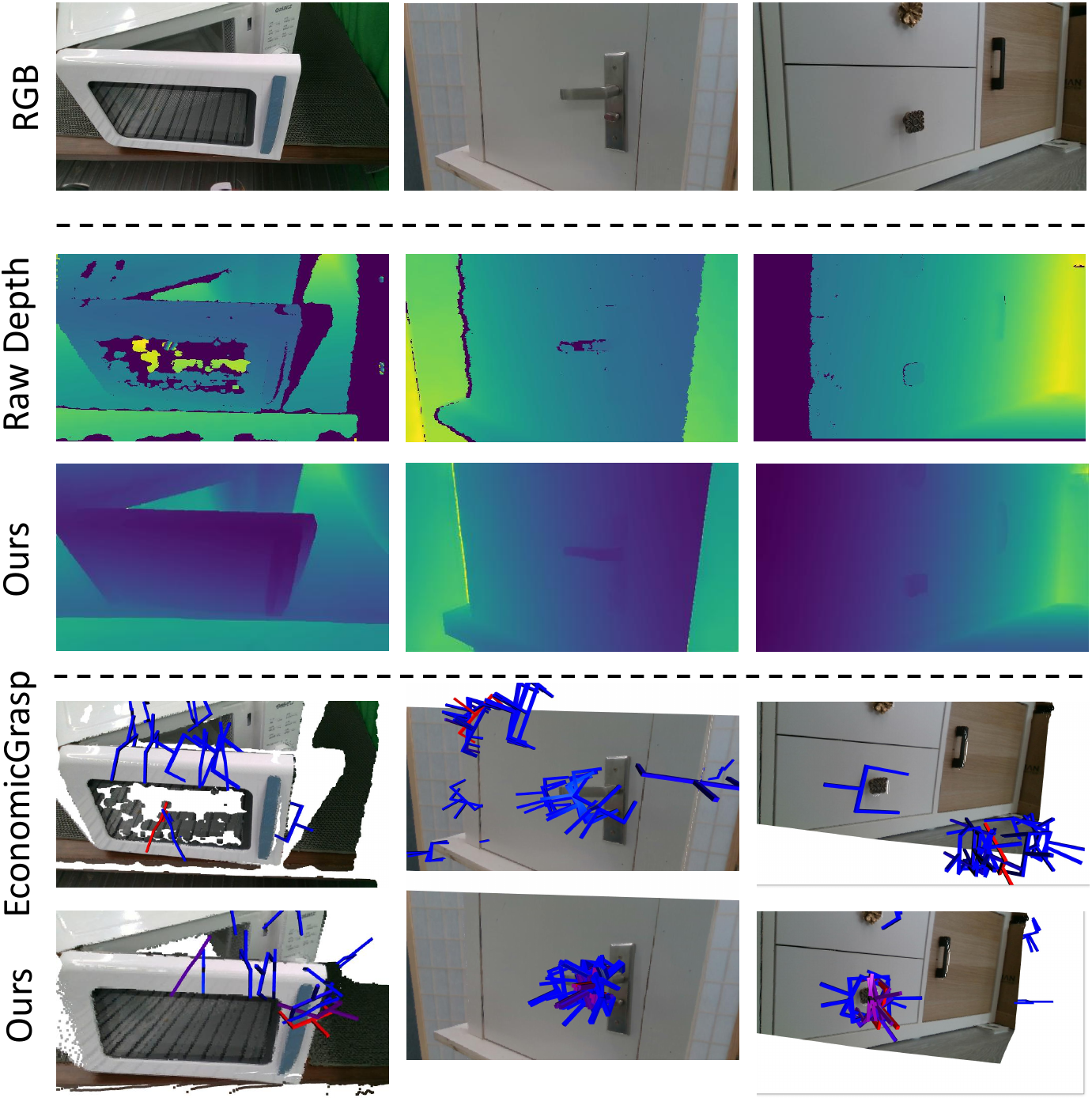}
    \caption{\textbf{Qualitative Results For Real-world Manipulation.} The actionable poses with top scores are displayed, with the red gripper representing the top-1 pose. }
    \label{fig:real_demo}
    % \vspace{-15pt}
\end{figure}

To validate the sim-to-real generalizability of our novel framework, we conduct real-world experiments. 
We use a Franka robot arm with an Intel RealSense camera to capture depth and IR images. 
We compare our method with three baselines: Where2Act, AO-Grasp, GSNet, and, like in Sec. \ref{sec:pose_experiments}, we extend the Where2Act interaction pipeline to finish our tasks. 
The experiments consist of 7 distinct instances, including StorageFurniture, Box, and Microwave. We evaluate the success rate of the top-1 interaction pose for each method across \textit{open} (n=14) and \textit{close} (n=17) tasks. 
As shown in Tab. \ref{table:real_world_experiment}, the overall success rate of our framework is 61.29\%, showcasing not only a successful transfer to the real world but also a significant performance boost compared to other methods.

Additionally, we perform ablation studies to assess how different modules affect the overall framework performance. As shown in Fig. \ref{fig:real_demo}, the depth camera yields poor depth maps when facing certain challenging materials, significantly impacting subsequent manipulation. 
Our depth reconstruction module effectively addresses this issue by recovering the depth map, thereby enhancing the performance of subsequent modules. 
Similarly, as illustrated in Fig. \ref{fig:real_demo}, Our framework, built upon our~\datasetname~dataset, tends to prioritize the interaction poses on actionable parts during prediction.
This part-aware capability could possibly explain why our method leads to such significant performance improvement shown in Tab. \ref{table:real_world_experiment}.

\begin{table}[ht]
\scriptsize
\centering
\begin{center}
\caption{\textbf{Quantitative Results for Real-world Articulated Object Manipulation}}
\begin{tabular}{lccc}
\bottomrule

\multicolumn{1}{c}{\multirow{2}{*}{Method}} & \multicolumn{3}{c}{Success Rate (\%) $\uparrow$ } \\ \cline{2-4} 
\multicolumn{1}{c}{}                        & Open & Close & Overall \\ \hline
AO-Grasp~\cite{morlans2023aograsp}                   & 28.57          & 29.41          & 29.03          \\
Where2Act~\cite{mo2021where2act}                  & 21.42          & 17.64          & 19.35          \\
GSNet~\cite{wang2021gsnet}         & 42.85          & 23.53         & 32.25          \\
 \hline
Ours w/o Part-aware EcoGrasp            & \textbf{64.28} & 41.17          & 51.61          \\
Ours w/o Depth Reconstruction             & 50.00          & 29.41          & 38.70          \\ 
Ours                       & \textbf{64.28} & \textbf{58.82} & \textbf{61.29} \\
\bottomrule
\end{tabular}
\end{center}
\label{table:real_world_experiment}
% \vspace{-15pt}
\end{table}

\section{Conclusions}
In this paper, we build a large-scale, part-centric synthetic dataset for material-agnostic articulated object manipulation. Our dataset is the first dataset featuring photo-realistic material randomization and part-oriented, scene-level actionable interaction pose annotation for articulated objects.
Building upon our dataset, we propose a novel articulated object manipulation framework capable of zero-shot transfer to the real world. 
We conduct experiments on individual modules and real-world overall experiments, with results indicating the competitiveness of our approach. Our dataset will be open-sourced.

{
\bibliographystyle{IEEEtran}
\bibliography{IEEEabrv,ref}

\begin{thebibliography}{10}
\providecommand{\url}[1]{#1}
\csname url@rmstyle\endcsname
\providecommand{\newblock}{\relax}
\providecommand{\bibinfo}[2]{#2}
\providecommand\BIBentrySTDinterwordspacing{\spaceskip=0pt\relax}
\providecommand\BIBentryALTinterwordstretchfactor{4}
\providecommand\BIBentryALTinterwordspacing{\spaceskip=\fontdimen2\font plus
\BIBentryALTinterwordstretchfactor\fontdimen3\font minus \fontdimen4\font\relax}
\providecommand\BIBforeignlanguage[2]{{%
\expandafter\ifx\csname l@#1\endcsname\relax
\typeout{** WARNING: IEEEtran.bst: No hyphenation pattern has been}%
\typeout{** loaded for the language `#1'. Using the pattern for}%
\typeout{** the default language instead.}%
\else
\language=\csname l@#1\endcsname
\fi
#2}}

\bibitem{Xiang_2020_SAPIEN}
F.~Xiang, Y.~Qin, K.~Mo, Y.~Xia, H.~Zhu, F.~Liu, M.~Liu, H.~Jiang, Y.~Yuan, H.~Wang, L.~Yi, A.~X. Chang, L.~J. Guibas, and H.~Su, ``{SAPIEN}: A simulated part-based interactive environment,'' in \emph{The IEEE Conference on Computer Vision and Pattern Recognition (CVPR)}, June 2020.

\bibitem{mo2021where2act}
K.~Mo, L.~J. Guibas, M.~Mukadam, A.~Gupta, and S.~Tulsiani, ``Where2act: From pixels to actions for articulated 3d objects,'' in \emph{Proceedings of the IEEE/CVF International Conference on Computer Vision}, 2021, pp. 6813--6823.

\bibitem{wu2021vatmart}
R.~Wu, Y.~Zhao, K.~Mo, Z.~Guo, Y.~Wang, T.~Wu, Q.~Fan, X.~Chen, L.~Guibas, and H.~Dong, ``Vat-mart: Learning visual action trajectory proposals for manipulating 3d articulated objects,'' \emph{arXiv preprint arXiv:2106.14440}, 2021.

\bibitem{wang2022adaafford}
Y.~Wang, R.~Wu, K.~Mo, J.~Ke, Q.~Fan, L.~J. Guibas, and H.~Dong, ``Adaafford: Learning to adapt manipulation affordance for 3d articulated objects via few-shot interactions,'' in \emph{European conference on computer vision}.\hskip 1em plus 0.5em minus 0.4em\relax Springer, 2022, pp. 90--107.

\bibitem{zhao2022dualafford}
Y.~Zhao, R.~Wu, Z.~Chen, Y.~Zhang, Q.~Fan, K.~Mo, and H.~Dong, ``Dualafford: Learning collaborative visual affordance for dual-gripper manipulation,'' \emph{arXiv preprint arXiv:2207.01971}, 2022.

\bibitem{eisner2022flowbot3d}
B.~Eisner, H.~Zhang, and D.~Held, ``Flowbot3d: Learning 3d articulation flow to manipulate articulated objects,'' \emph{arXiv preprint arXiv:2205.04382}, 2022.

\bibitem{zhang2023flowbot++}
H.~Zhang, B.~Eisner, and D.~Held, ``Flowbot++: Learning generalized articulated objects manipulation via articulation projection,'' \emph{arXiv preprint arXiv:2306.12893}, 2023.

\bibitem{zhong20233d}
C.~Zhong, Y.~Zheng, Y.~Zheng, H.~Zhao, L.~Yi, X.~Mu, L.~Wang, P.~Li, G.~Zhou, C.~Yang, \emph{et~al.}, ``3d implicit transporter for temporally consistent keypoint discovery,'' in \emph{Proceedings of the IEEE/CVF International Conference on Computer Vision}, 2023, pp. 3869--3880.

\bibitem{geng2023gapartnet}
H.~Geng, H.~Xu, C.~Zhao, C.~Xu, L.~Yi, S.~Huang, and H.~Wang, ``Gapartnet: Cross-category domain-generalizable object perception and manipulation via generalizable and actionable parts,'' in \emph{Proceedings of the IEEE/CVF Conference on Computer Vision and Pattern Recognition}, 2023, pp. 7081--7091.

\bibitem{geng2023partmanip}
H.~Geng, Z.~Li, Y.~Geng, J.~Chen, H.~Dong, and H.~Wang, ``Partmanip: Learning cross-category generalizable part manipulation policy from point cloud observations,'' in \emph{Proceedings of the IEEE/CVF Conference on Computer Vision and Pattern Recognition}, 2023, pp. 2978--2988.

\bibitem{geng2024sage}
H.~Geng, S.~Wei, C.~Deng, B.~Shen, H.~Wang, and L.~Guibas, ``Sage: Bridging semantic and actionable parts for generalizable manipulation of articulated objects,'' 2024.

\bibitem{wang2024rpmart}
J.~Wang, W.~Liu, Q.~Yu, Y.~You, L.~Liu, W.~Wang, and C.~Lu, ``Rpmart: Towards robust perception and manipulation for articulated objects,'' \emph{arXiv preprint arXiv:2403.16023}, 2024.

\bibitem{rlafford}
Y.~Geng, B.~An, H.~Geng, Y.~Chen, Y.~Yang, and H.~Dong, ``Rlafford: End-to-end affordance learning for robotic manipulation,'' in \emph{2023 IEEE International Conference on Robotics and Automation (ICRA)}, 2023, pp. 5880--5886.

\bibitem{wei2024d3roma}
S.~Wei, H.~Geng, J.~Chen, C.~Deng, W.~Cui, C.~Zhao, X.~Fang, L.~Guibas, and H.~Wang, ``D3roma: Disparity diffusion-based depth sensing for material-agnostic robotic manipulation,'' in \emph{8th Annual Conference on Robot Learning (CoRL)}, 2024.

\bibitem{shi2024asgrasp}
J.~Shi, A.~Yong, Y.~Jin, D.~Li, H.~Niu, Z.~Jin, and H.~Wang, ``Asgrasp: Generalizable transparent object reconstruction and 6-dof grasp detection from rgb-d active stereo camera,'' in \emph{2024 IEEE International Conference on Robotics and Automation (ICRA)}.\hskip 1em plus 0.5em minus 0.4em\relax IEEE, 2024, pp. 5441--5447.

\bibitem{morlans2023aograsp}
C.~P. Morlans, C.~Chen, Y.~Weng, M.~Yi, Y.~Huang, N.~Heppert, L.~Zhou, L.~Guibas, and J.~Bohg, ``Ao-grasp: Articulated object grasp generation,'' \emph{arXiv preprint arXiv:2310.15928}, 2023.

\bibitem{an2024rgbmanip}
B.~An, Y.~Geng, K.~Chen, X.~Li, Q.~Dou, and H.~Dong, ``Rgbmanip: Monocular image-based robotic manipulation through active object pose estimation,'' in \emph{2024 IEEE International Conference on Robotics and Automation (ICRA)}.\hskip 1em plus 0.5em minus 0.4em\relax IEEE, 2024, pp. 7748--7755.

\bibitem{yi2018deep}
L.~Yi, H.~Huang, D.~Liu, E.~Kalogerakis, H.~Su, and L.~Guibas, ``Deep part induction from articulated object pairs,'' \emph{arXiv preprint arXiv:1809.07417}, 2018.

\bibitem{deng2024banana}
C.~Deng, J.~Lei, W.~B. Shen, K.~Daniilidis, and L.~J. Guibas, ``Banana: Banach fixed-point network for pointcloud segmentation with inter-part equivariance,'' in \emph{NeurIPS}, 2024.

\bibitem{li2020category}
X.~Li, H.~Wang, L.~Yi, L.~J. Guibas, A.~L. Abbott, and S.~Song, ``Category-level articulated object pose estimation,'' in \emph{CVPR}, 2020.

\bibitem{liu2023semi}
G.~Liu, Q.~Sun, H.~Huang, C.~Ma, Y.~Guo, L.~Yi, H.~Huang, and R.~Hu, ``Semi-weakly supervised object kinematic motion prediction,'' in \emph{CVPR}, 2023.

\bibitem{lyu2024scissorbot}
\BIBentryALTinterwordspacing
J.~Lyu, Y.~Chen, T.~Du, F.~Zhu, H.~Liu, Y.~Wang, and H.~Wang, ``Scissorbot: Learning generalizable scissor skill for paper cutting via simulation, imitation, and sim2real,'' in \emph{8th Annual Conference on Robot Learning}, 2024. [Online]. Available: \url{https://openreview.net/forum?id=PAtsxVz0ND}
\BIBentrySTDinterwordspacing

\bibitem{zhang2024gamma}
J.~Zhang, N.~Gireesh, J.~Wang, X.~Fang, C.~Xu, W.~Chen, L.~Dai, and H.~Wang, ``Gamma: Graspability-aware mobile manipulation policy learning based on online grasping pose fusion,'' in \emph{2024 IEEE International Conference on Robotics and Automation (ICRA)}.\hskip 1em plus 0.5em minus 0.4em\relax IEEE, 2024, pp. 1399--1405.

\bibitem{chen2023urdformer}
Q.~Chen, M.~Memmel, A.~Fang, A.~Walsman, D.~Fox, and A.~Gupta, ``Urdformer: Constructing interactive realistic scenes from real images via simulation and generative modeling,'' in \emph{Towards Generalist Robots: Learning Paradigms for Scalable Skill Acquisition @ CoRL 2023}, 2023.

\bibitem{mu2021sdf}
J.~Mu, W.~Qiu, A.~Kortylewski, A.~Yuille, N.~Vasconcelos, and X.~Wang, ``A-sdf: Learning disentangled signed distance functions for articulated shape representation,'' in \emph{ICCV}, 2021.

\bibitem{jiang2022ditto}
Z.~Jiang, C.-C. Hsu, and Y.~Zhu, ``Ditto: Building digital twins of articulated objects from interaction,'' in \emph{CVPR}, 2022.

\bibitem{tseng2022cla}
W.-C. Tseng, H.-J. Liao, L.~Yen-Chen, and M.~Sun, ``Cla-nerf: Category-level articulated neural radiance field,'' in \emph{ICRA}, 2022.

\bibitem{luo2024physpartphysicallyplausiblecompletion}
\BIBentryALTinterwordspacing
R.~Luo, H.~Geng, C.~Deng, P.~Li, Z.~Wang, B.~Jia, L.~Guibas, and S.~Huang, ``Physpart: Physically plausible part completion for interactable objects,'' 2024. [Online]. Available: \url{https://arxiv.org/abs/2408.13724}
\BIBentrySTDinterwordspacing

\bibitem{lei2023nap}
J.~Lei, C.~Deng, B.~Shen, L.~Guibas, and K.~Daniilidis, ``Nap: Neural 3d articulation prior,'' \emph{arXiv preprint arXiv:2305.16315}, 2023.

\bibitem{liu2024cage}
J.~Liu, H.~I.~I. Tam, A.~Mahdavi-Amiri, and M.~Savva, ``Cage: Controllable articulation generation,'' in \emph{CVPR}, 2024.

\bibitem{geng2022end}
Y.~Geng, B.~An, H.~Geng, Y.~Chen, Y.~Yang, and H.~Dong, ``End-to-end affordance learning for robotic manipulation,'' in \emph{ICRA}, 2023.

\bibitem{gong2023arnold}
R.~Gong, J.~Huang, Y.~Zhao, H.~Geng, X.~Gao, Q.~Wu, W.~Ai, Z.~Zhou, D.~Terzopoulos, S.-C. Zhu, \emph{et~al.}, ``Arnold: A benchmark for language-grounded task learning with continuous states in realistic 3d scenes,'' in \emph{ICCV}, 2023.

\bibitem{fang2020graspnet}
H.-S. Fang, C.~Wang, M.~Gou, and C.~Lu, ``Graspnet-1billion: A large-scale benchmark for general object grasping,'' in \emph{Proceedings of the IEEE/CVF Conference on Computer Vision and Pattern Recognition}, 2020, pp. 11\,444--11\,453.

\bibitem{sundermeyer2021contactgraspnetefficient6dofgrasp}
\BIBentryALTinterwordspacing
M.~Sundermeyer, A.~Mousavian, R.~Triebel, and D.~Fox, ``Contact-graspnet: Efficient 6-dof grasp generation in cluttered scenes,'' 2021. [Online]. Available: \url{https://arxiv.org/abs/2103.14127}
\BIBentrySTDinterwordspacing

\bibitem{guhur2023instruction}
P.-L. Guhur, S.~Chen, R.~G. Pinel, M.~Tapaswi, I.~Laptev, and C.~Schmid, ``Instruction-driven history-aware policies for robotic manipulations,'' in \emph{Conference on Robot Learning}.\hskip 1em plus 0.5em minus 0.4em\relax PMLR, 2023, pp. 175--187.

\bibitem{liu2024composablepartbasedmanipulation}
\BIBentryALTinterwordspacing
W.~Liu, J.~Mao, J.~Hsu, T.~Hermans, A.~Garg, and J.~Wu, ``Composable part-based manipulation,'' 2024. [Online]. Available: \url{https://arxiv.org/abs/2405.05876}
\BIBentrySTDinterwordspacing

\bibitem{ling2024articulatedobjectmanipulationcoarsetofine}
\BIBentryALTinterwordspacing
S.~Ling, Y.~Wang, S.~Wu, Y.~Zhuang, T.~Xu, Y.~Li, C.~Liu, and H.~Dong, ``Articulated object manipulation with coarse-to-fine affordance for mitigating the effect of point cloud noise,'' 2024. [Online]. Available: \url{https://arxiv.org/abs/2402.18699}
\BIBentrySTDinterwordspacing

\bibitem{isaac-sim}
J.~Liang, V.~Makoviychuk, A.~Handa, N.~Chentanez, M.~Macklin, and D.~Fox, ``Gpu-accelerated robotic simulation for distributed reinforcement learning,'' 2018.

\bibitem{dai2022dreds}
Q.~Dai, J.~Zhang, Q.~Li, T.~Wu, H.~Dong, Z.~Liu, P.~Tan, and H.~Wang, ``Domain randomization-enhanced depth simulation and restoration for perceiving and grasping specular and transparent objects,'' in \emph{European Conference on Computer Vision (ECCV)}, 2022.

\bibitem{wu2024economicframework6dofgrasp}
\BIBentryALTinterwordspacing
X.-M. Wu, J.-F. Cai, J.-J. Jiang, D.~Zheng, Y.-L. Wei, and W.-S. Zheng, ``An economic framework for 6-dof grasp detection,'' 2024. [Online]. Available: \url{https://arxiv.org/abs/2407.08366}
\BIBentrySTDinterwordspacing

\bibitem{curobo}
B.~Sundaralingam, S.~K.~S. Hari, A.~Fishman, C.~Garrett, K.~Van~Wyk, V.~Blukis, A.~Millane, H.~Oleynikova, A.~Handa, F.~Ramos, N.~Ratliff, and D.~Fox, ``Curobo: Parallelized collision-free robot motion generation,'' in \emph{2023 IEEE International Conference on Robotics and Automation (ICRA)}, 2023, pp. 8112--8119.

\bibitem{hirschmuller2008sgm}
H.~Hirschmuller, ``Stereo processing by semiglobal matching and mutual information,'' \emph{IEEE Transactions on Pattern Analysis and Machine Intelligence}, vol.~30, no.~2, pp. 328--341, 2008.

\bibitem{lipson2021raftstereo}
L.~Lipson, Z.~Teed, and J.~Deng, ``Raft-stereo: Multilevel recurrent field transforms for stereo matching,'' in \emph{2021 International Conference on 3D Vision (3DV)}.\hskip 1em plus 0.5em minus 0.4em\relax IEEE, 2021, pp. 218--227.

\bibitem{teed2020raft}
Z.~Teed and J.~Deng, ``Raft: Recurrent all-pairs field transforms for optical flow,'' in \emph{Computer Vision--ECCV 2020: 16th European Conference, Glasgow, UK, August 23--28, 2020, Proceedings, Part II 16}.\hskip 1em plus 0.5em minus 0.4em\relax Springer, 2020, pp. 402--419.

\bibitem{wang2021gsnet}
C.~Wang, H.-S. Fang, M.~Gou, H.~Fang, J.~Gao, and C.~Lu, ``Graspness discovery in clutters for fast and accurate grasp detection,'' in \emph{Proceedings of the IEEE/CVF International Conference on Computer Vision}, 2021, pp. 15\,964--15\,973.

\end{thebibliography}
}

\end{document}